\RequirePackage{snapshot}
\documentclass{article}

\usepackage{microtype}
\usepackage{graphicx}
\usepackage{subcaption}
\usepackage{booktabs}
\usepackage{enumitem}
\usepackage{hyperref}

\usepackage[accepted]{icml2023}

\usepackage{amsmath}
\usepackage{amssymb}
\usepackage{mathtools}
\usepackage{amsthm}

\usepackage[capitalize,noabbrev]{cleveref}

\theoremstyle{plain}

\theoremstyle{definition}

\theoremstyle{remark}

\usepackage[textsize=tiny]{todonotes}
\usepackage{todonotes}

\newcommand{\Fixme}[2][]{\noindent}
\newcommand{\Notewho}[3][]{\noindent}
\newcommand{\Chenghao}[2][]{\noindent}
\newcommand{\Ziy}[2][]{\noindent}

\usepackage{csquotes}

\usepackage{array}
\newcolumntype{H}{>{\setbox0=\hbox\bgroup}c<{\egroup}@{}}

\icmltitlerunning{Equipping Transformer with Random-Access Reading for Long-Context Understanding}

\begin{document}

\twocolumn[
\icmltitle{Equipping Transformer with Random-Access Reading \\ for Long-Context Understanding \\
           }

\icmlsetsymbol{equal}{*}

\begin{icmlauthorlist}
\icmlauthor{Chenghao Yang}{equal,uchicago}
\icmlauthor{Zi Yang}{equal,google}
\icmlauthor{Nan Hua}{google}
\end{icmlauthorlist}

\icmlaffiliation{uchicago}{University of Chicago (Work is done at Google as a student researcher)}
\icmlaffiliation{google}{Google Research}

\icmlcorrespondingauthor{Chenghao Yang}{yangalan1996@gmail.com}
\icmlcorrespondingauthor{Zi Yang}{ziy@google.com}

\icmlkeywords{Machine Learning, ICML}

\vskip 0.3in
]

\printAffiliationsAndNotice{\icmlEqualContribution}

\begin{abstract}
Long-context modeling presents a significant challenge for transformer-based large language models (LLMs) due to the quadratic complexity of the self-attention mechanism and issues with length extrapolation caused by pretraining exclusively on short inputs. Existing methods address computational complexity through techniques such as text chunking, the kernel approach, and structured attention, and tackle length extrapolation problems through positional encoding, continued pretraining, and data engineering. These approaches typically require \textbf{sequential access} to the document, necessitating reading from the first to the last token. We contend that for goal-oriented reading of long documents, such sequential access is not necessary, and a proficiently trained model can learn to omit hundreds of less pertinent tokens.
Inspired by human reading behaviors and existing empirical observations, we propose \textbf{random access}, a novel reading strategy that enables transformers to efficiently process long documents without examining every token. Experimental results from pretraining, fine-tuning, and inference phases validate the efficacy of our method.
\end{abstract}

\section{Introduction}
\label{sec:intro}
Long context refers to an input context that exceeds the maximum length limit,\footnote{We acknowledge the recent achievement building models that can handle  extremely long context, such as Gemini 1.5~\cite{reid2024gemini}. However, it is impossible to expand the context window indefinitely without chunking it at some point, and we believe combining chunking with our proposed random-access reading strategy can be more efficient than directly reading the input as a whole.} making it impossible to process in a single inference step. Examples of long context inputs include multiple webpages~\cite{zhou2023webarena, deng2024mind2web}, books~\cite{mou2021narrative}, code repository~\cite{jimenez2023swe} and dialog histories~\cite{yang2023can}.

Existing strategies predominantly employ a \emph{sequential-access} model, where both during training and inference, the model processes tokens of documents in their original, sequential order~\cite{dong2023survey, huang2023advancing}.
These methods utilize various mechanisms to address challenges such as quadratic complexity—through block-wise processing~\cite{qiu2020blockwise, tay2020sparse, liu2022leveraging, ivgi2023efficient, mohtashami2024random}, structured attention~\cite{Beltagy2020Longformer, guo2022longt5, xiao2023efficient, han2023lm}, and linear approximations~\cite{choromanski2020rethinking, peng2020random, ma2021luna, nguyen2022fourierformer}—or to mitigate length extrapolation issues using techniques like rotary positional embeddings~\cite{peng2023yarn,su2024roformer} and continual training~\cite{xiong2023effective,fu2024data}.

While these methods have been widely adopted and proven effective, they treat every token as equally important and overlook the fact that for user queries, only a small portion of the information is relevant~\cite{ding2020cogltx}. Consequently, many proposed solutions still suffer from computational overhead, particularly in online interactions between humans and LLMs. Recent works in retrieval-augmented generation (RAG)~\cite{lewis2020retrieval, shi2023replug} have attempted to address this by incorporating an additional retriever to bypass non-essential context and selectively retrieve relevant sections. However, due to an incomplete understanding of the entire long context and a lack of robust supervisory signals, the overall performance of such multi-module systems is significantly constrained by the capabilities of the retriever~\cite{mou2021narrative, zhang2022summn}. Furthermore, while these RAG systems require a predefined top-K for any user query, our framework dynamically determines the access pattern based on the query and the context.

Inspired by strategies observed in proficient human readers, who actively engage with long texts by developing predictions of forthcoming content and selectively skip-reading irrelevant sections~\cite{paris1991development, pressley2012verbal}, along with recent evidence suggesting that large models may inherently acquire the ability to access content at arbitrary locations during pretraining~\cite{fu2024data}, we propose the \emph{Random-Access Reading}.\footnote{Random access is typically defined as the capability to access arbitrary positions, as in a Random-Access Machine~\cite{knuth1997art}. Within the scope of this paper, we focus on a simplified version where we only allow the model to access arbitrary positions in the future context and never look back, to achieve better efficiency.} This approach leverages local perplexity information as a criterion to bypass non-essential text, thereby significantly enhancing reading efficiency.
\begin{figure}[hpt!]
    \centering
    \includegraphics[width=\columnwidth, trim=2cm 3cm 3cm 3cm, clip]{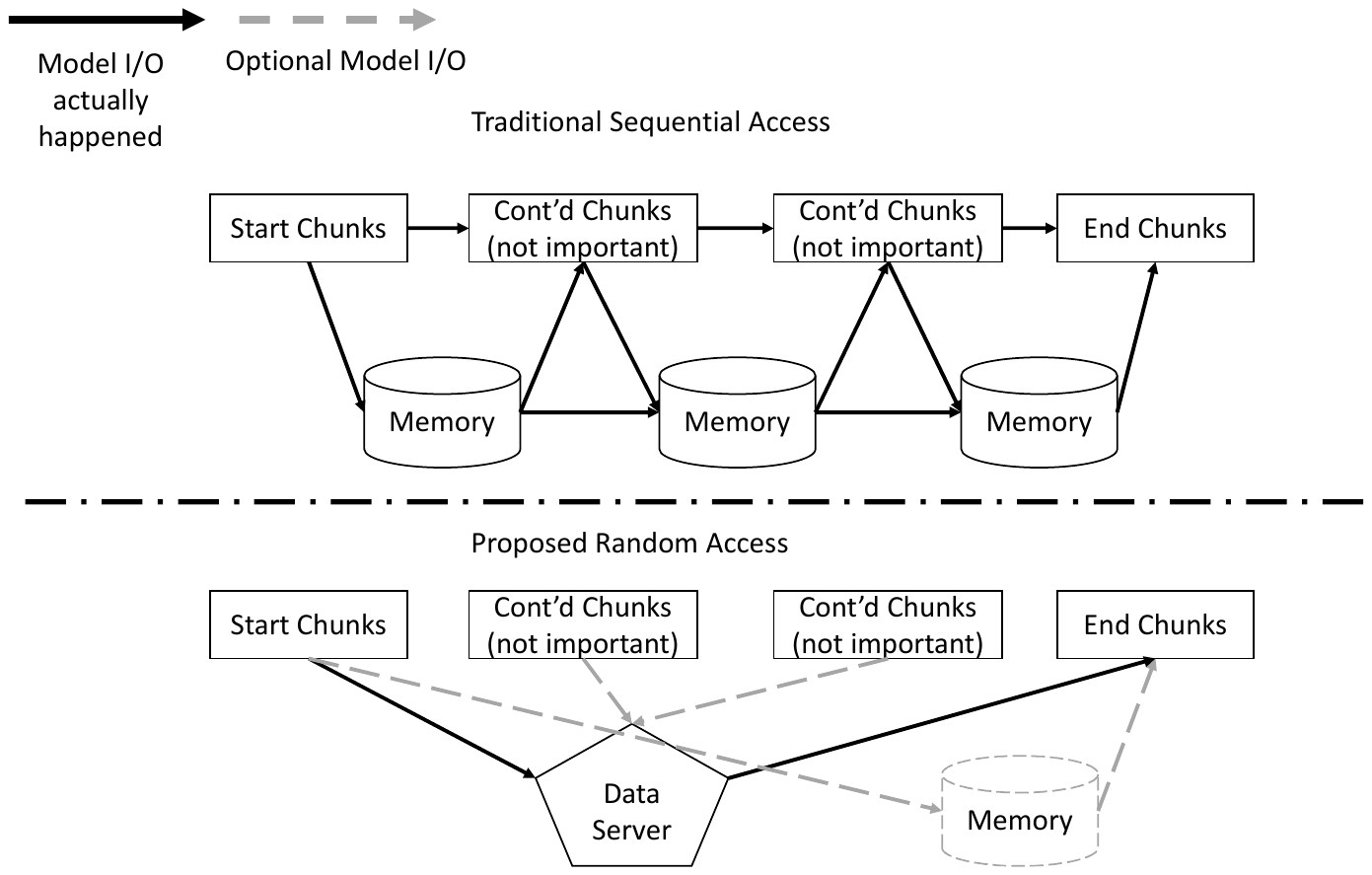}
    \caption{Illustration for our proposed random-access reading strategy for long-context modeling. In the traditional sequential access scenario (upper part of the figure), inputs are split into equal-sized chunks and fed to the model in sequential order. In contrast, we propose building an additional data server module (in the lower part of the figure) that takes relevant statistics from the model and decides which chunk it should read next. }
    \label{fig:architecture}
    \end{figure}

The proposed architecture is depicted in \cref{fig:architecture}. Unlike traditional I/O access, where each chunk of input is fed sequentially to the model, we propose building a new \textit{data server} to handle the I/O process. The random access we propose involves the model transmitting relevant statistics\footnote{In this paper, we demonstrate the effectiveness of random-access reading using pooled per-token cross entropy loss as such statistics, although future research could explore alternative metrics.} to the data server, which then employs our specially designed skipping mechanism (outlined in \cref{sec: skipping_method}) to determine the number of tokens to be skipped and subsequently fetch a new chunk for the model. Optionally, we can use a memory module\footnote{This will make our model similar to the landmark attention mechanism proposed by \cite{mohtashami2023landmark}. We will elaborate on the differences between our work and theirs in \cref{sec: background} and discuss memory module implementation in \cref{sec: skipping_method}.} to provide additional context, helping the model make better skipping decisions and maintain a coherent understanding of the input text.

Through extensive experiments, we have validated the effectiveness of our proposed method. Our findings include:
\begin{enumerate}[leftmargin=*]
\item The skipping mechanism enhances model performance in long-context language model pretraining.
\item Traditional short-text pretraining degrades model performance below that of a randomly-initialized model. However, with our fine-tuning using skipping mechanism, a short-text model can be successfully adapted to excel in long-context tasks, outperforming even our random-access model pretrained from scratch.
\item Incorporating a memory module allows our random-access model to achieve further significant improvements, surpassing the previous state-of-the-art memory-based models in pretraining~\cite{wu2021memorizing} with only 26$\%$ of the training time.
\item Evaluating our random-access model on downstream tasks using a modified TriviaQA~\cite{joshi2017triviaqa} task, where all retrieved evidence from the original dataset is concatenated to create an extremely challenging long-context scenario—demonstrates that more aggressive skipping mechanism yield better performance. This confirms the method's effectiveness and improved learning efficiency.
\end{enumerate}

\section{Background}
\label{sec: background}
When applied to Transformer models, our proposed random-access reading strategies can significantly reduce attention computations by skipping many tokens. This aligns closely with research on attention sparsification that uses local information to enhance long-context modeling efficiency and capability. Notably, \citet{liu2023deja} demonstrate the existence of contextual sparsity that can be leveraged for computational acceleration. \citet{chowdhury2023monotonic} illustrate the effectiveness of using sliding windows and weighted decay for long-context understanding tasks. Additionally, \citet{fu2023simple} find that a simple convolutional model performs well on Long Range Arena tasks~\cite{tay2020long}. \citet{chen2023walking} develop a bottom-up hierarchical structure that utilizes an LLM to read long documents, enabling direct navigation to document sections relevant to specific queries by following a tree path. Further, \citet{han2023lm} and \citet{xiao2023efficient} report that maintaining attention on only a few initial tokens (termed "attention sinks"), with each token attending to just its neighbors, can significantly enhance the efficiency of processing long contexts while preserving performance. These works provides both empirical and theoretical support for conducting attention sparsification, and our random-access transformer takes a more aggressive approach by directly skipping many tokens.

The most closely related concept to our random-access model is the landmark attention mechanism introduced by \citet{mohtashami2023landmark}. In this approach, a landmark token is appended to the end of each fixed-length chunk to represent that chunk, and each token attends only to a limited number of these chunk-wise landmark tokens. While this method facilitates random access for each token and significantly reduces memory and computational requirements, it still necessitates choosing a constant chunk size—which may vary across tasks—reads the entire context sequentially (thus retaining quadratic complexity albeit reduced by a constant factor), and requires implementing specific data structures and approximate retrieval algorithms for efficiency. In contrast, our method leverages aggregated statistics to directly skip to the next informative window. During each reading phase, our model fully utilizes the context window for which it was trained, rather than relying on a manually specified chunk size, and optionally incorporates a memory module (aligning it closer to the "attention sink" approach). This allows for better modeling of text coherence and achieves near sublinear complexity.\footnote{We exclude the computational complexity of the tokenization process as it is relatively lightweight. In fact, to omit tokenization, our approach can also work with a character-based Transformer model out of the box, where the semantics of the offset now automatically means characters rather than sentencepiece.}

\section{Skipping Mechanism}
\label{sec: skipping_method}
When reading a large section of a document, we posit that even a half-way trained language model might already have a reasonable prediction for some future segments, which can therefore be safely skipped. The skipping mechanism proposed is straightforward: Suppose model $M$ begins reading at the $S$-th token, $X_S$, of document $X$, and continues for $L(M)$ tokens—typically the maximum token limit for the model, e.g., $L(M) = 512$. The reading ends at the token $X_{S+L(M)}$. During this interval, the model performs self-attention over the span ${X_S, \dots, X_{S+L(M)}}$ and computes token-wise cross entropy losses ${\mathcal{L}_S, \dots, \mathcal{L}_{S+L(M)}}$. We then apply a pooling operation to these losses to estimate the model's confidence over this passage, denoted as $C(X, S; M) = \text{pooling}(\mathcal{L}_S, \dots, \mathcal{L}_{S+L(M)})$. This confidence metric is used to determine the number of tokens, $D(X, S; M)$, that the model will subsequently skip:

\begin{align}
    &D(X, S; M) = \\ \nonumber
    &K \min \left\{\lfloor\frac{|X| - S - L(M)}{K}\rfloor, \lfloor \frac{\alpha}{C(X, S; M)} \rfloor \right\}
\end{align}

At the next reading step, the model resumes reading from the token $X_{S+L(M) + D(X, S; M)}$, and this process of skipping continues until the end of the document. Here, $K$ represents the skipping rate and $\alpha$ the skipping threshold. When $C(X, S; M)$ is small relative to $\alpha$, it suggests that the model has a robust understanding of the current passage, allowing it to safely skip at a rate of $K \lfloor \frac{\alpha}{C(X, S; M)} \rfloor$.

Intuitively, for pretraining and finetuning, where the loss is computed in a teacher-forcing paradigm, $C(X, S; M)$ directly measure how well the model can predict the next gold tokens given the previous context. The larger $C(X, S; M)$ is, the less confidence that a half-way trained model have for future context and thus should skip in a more conservative rate. During inference, without gold supervision, the situation becomes more challenging. However, prior works~\cite{dong2018confidence, kamath2020selective, jiang2021can} have identified  log-probability loss as a common measurement of model confidence at inference time, which we anticipate could be highly informative for making skipping decisions.

The simplicity of this heuristic underscores its value: it eliminates the need for additional models to predict the number of tokens to skip. Since the required per-token loss is a direct output of any standard language model, this mechanism can be seamlessly integrated into the normal language model pretraining process. Furthermore, our approach does not rely on any structural assumptions or intermediate representations; therefore, the skipping operations do not interfere with ongoing model operations and are compatible with any model structure that provides a probability output for each token. For simplicity, within the scope of this paper, we utilize the widely-recognized auto-regressive Transformer model.

Our experiments demonstrate that local skipping improves the efficiency-diversity trade-off in long-context modeling. Moreover, skipping-based fine-tuning enables an existing language model checkpoint to effectively handle long-context scenarios, surpassing even specialized memory-based baselines.

\paragraph{Working with Memory Mechanism}
Given the inherently limited context lengths of current language models and the irreversible nature of our skipping mechanism, there is a potential for making skipping decisions without sufficient context. Moreover, previous research indicates that maintaining attention over initial and neighboring tokens is crucial for understanding extended contexts~\cite{han2023lm, xiao2023efficient}; thus, focusing on these tokens can significantly enhance performance. To augment the model’s capacity for context processing and informed skipping, we introduce an optional memory mechanism into our framework. In this study, we implement the Attendre model, as proposed in \cite{yang2024attendre}, which utilizes a First-In First-Out (FIFO) memory eviction strategy alongside an approximated Top-K Key-Value retrieval algorithm. Unlike traditional memory-augmented transformer models such as the Memorizing Transformer \cite{wu2021memorizing}, our approach features a global memory pool that performs Top-K Key-Value retrieval across both the current reading window and past memory items at every layer, rather than being confined to a single intermediate layer.

\paragraph{Extension to Structured Documents}
Our approach can be further strengthened by leveraging hierarchical structures (e.g., the DOM tree for a web page or the table of contents in a novel). We can restrict skipping to occur within subtrees, perform simultaneous skipping for multiple parts of the document, and aggregate information globally. This method aligns more closely with the landmark attention approach~\cite{mohtashami2023landmark}, featuring uneven block sizes and more coherent block semantics. Even without a prebuilt index, recursive summarization can be employed to build such an index from scratch, as demonstrated in \cite{chen2023walking}. In this work, our focus is on showcasing the effectiveness of our proposed skipping mechanism without assuming any structure in the input document. Therefore, we leave further exploration in this area for future work.

\section{Long-Context Language Modeling}
\label{sec: long-context lm}
We first evaluate the effectiveness of our random-access method in pre-training over a long-context language modeling task—specifically, predicting the next token in a long discourse. We selected the C4 corpus~\cite{raffel2020exploring} for our task and filtered out documents with fewer than 4,000 tokens, resulting in the same corpus subset C4~(4k+) as used in \citep{wu2021memorizing}. To verify the effectiveness of our proposed skipping mechanism, we investigate two application scenarios: 1) \textbf{Pretraining}: Assuming sufficient computational resources are available, we aim to pretrain a model from scratch that can manage long contexts, where using skipping as a data feeding strategy could enhance model performance. 2) \textbf{Finetuning}: In practical situations where pretraining is not feasible, we consider whether the skipping mechanism can serve as a finetuning strategy to augment the capability of a short-context model in handling long contexts.

\begin{figure}[hpt!]
    \centering
    \includegraphics[width=0.93\columnwidth, trim=0cm 0cm 0cm 0cm, clip]{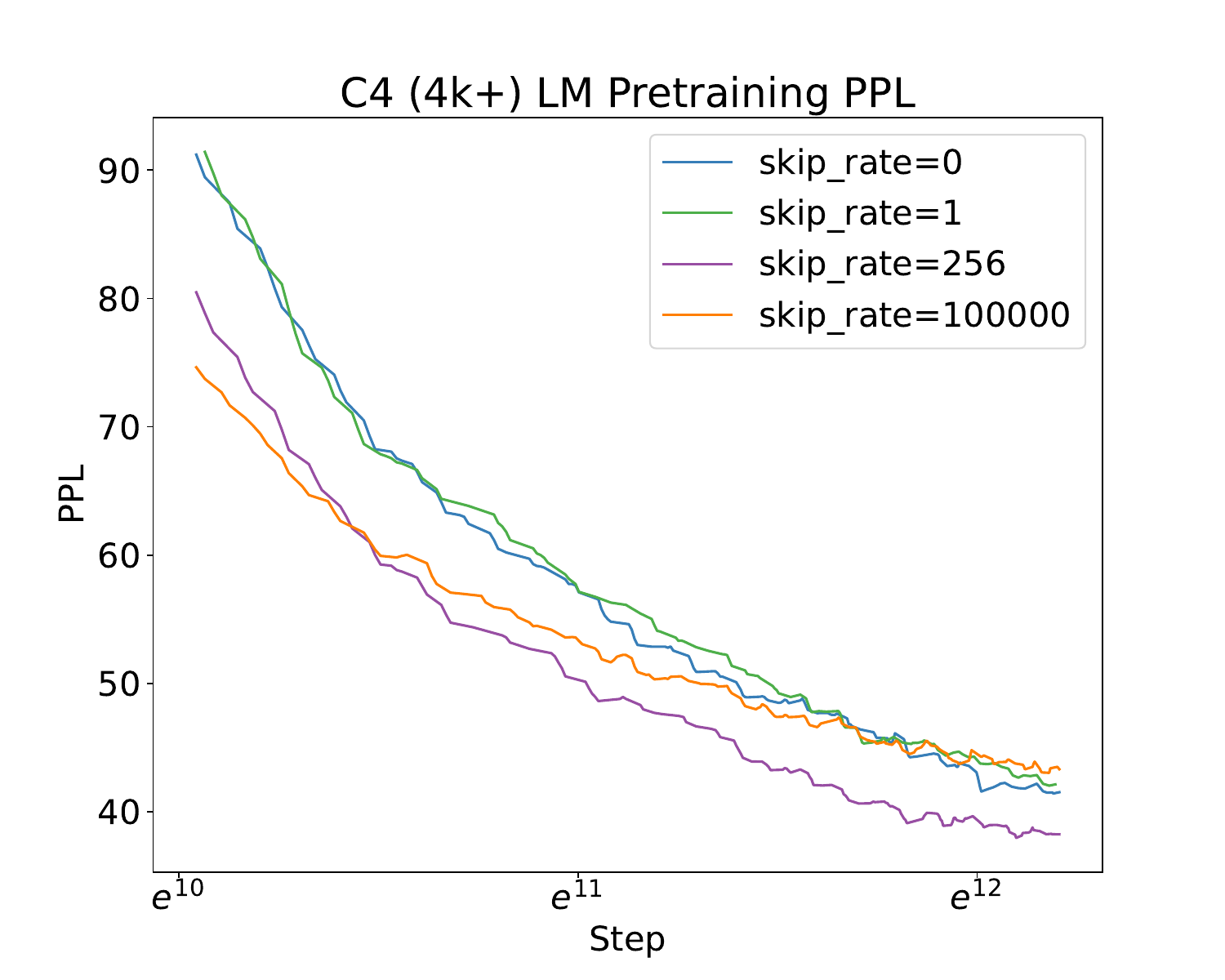}
    \vspace{-5mm}
    \caption{Pretraining Experiments on C4~(4k+) for models w/o memory mechanism. We disable skipping in evaluation time for fair comparison. Intermediate skipping rate performs best after sufficient training, which confirms the effectiveness of our method. }
    \label{fig:pretraining_wo_memory}
    \end{figure}

\subsection{Task Setup} 
\paragraph{Pretraining} We pretrain a 12-layer transformer model over $200,000$ steps with a batch size of $128$ and maximum token limit $L(M)=512$. In this experiment, we do not employ a memory mechanism to isolate and directly assess the effectiveness of the skipping mechanism. We set varying skipping rates $K=0, 1, 256, 100,000$ for different runs. For evaluation, we disable the skipping to ensure a fair comparison.

\paragraph{Finetuning} We utilize a language model pretrained on the C4 (short-text) corpus, specifically for next-word prediction over $100M$ steps, without the use of a skipping mechanism. We use the same model architecture as in pretraining task. In this short-text pretraining process, we follow the common data processing strategy (``random chunking and shuffling'') to concatenate all documents, split the whole corpus to $L(M)$-sized chunks and random shuffling the chunks~\cite{devlin-etal-2019-bert, raffel2020exploring}.
The finetuning is then conducted over $25,000$ steps on the C4~(4k+) training set used in the pretraining Task. We report the skipping rates $K=0, 256$ for simplicity.

\paragraph{Metric} We adopt perplexity (PPL) as the metric as in previous works for long-context language modeling. 

\subsection{Experiment Results}
\paragraph{Skipping achieves better long-context pretraining results.} The pretraining results are illustrated in \cref{fig:pretraining_wo_memory}. With sufficient training duration, an intermediate skip-rate (K=$256$) significantly reduces perplexity compared to non-skipping (K=$0$), demonstrating the effectiveness of the skipping mechanism in long-context pretraining. At very high skipping rates (i.e., K=$100,000$), the correlation between consecutively read documents diminishes, closely approximating the effect of input random shuffling. We find that skipping at a moderate rate outperforms both extremely fast skipping and non-skipping strategies. Consequently, our skipping mechanism not only maintains better coherence between chunked passages but also enhances model exposure to unique tokens, thereby improving generalization compared to consecutive chunked and random shuffling pretraining strategies.

\begin{figure}[]
\small
    \centering
    \includegraphics[width=0.93\columnwidth, trim=0cm 0cm 0cm 0cm, clip]{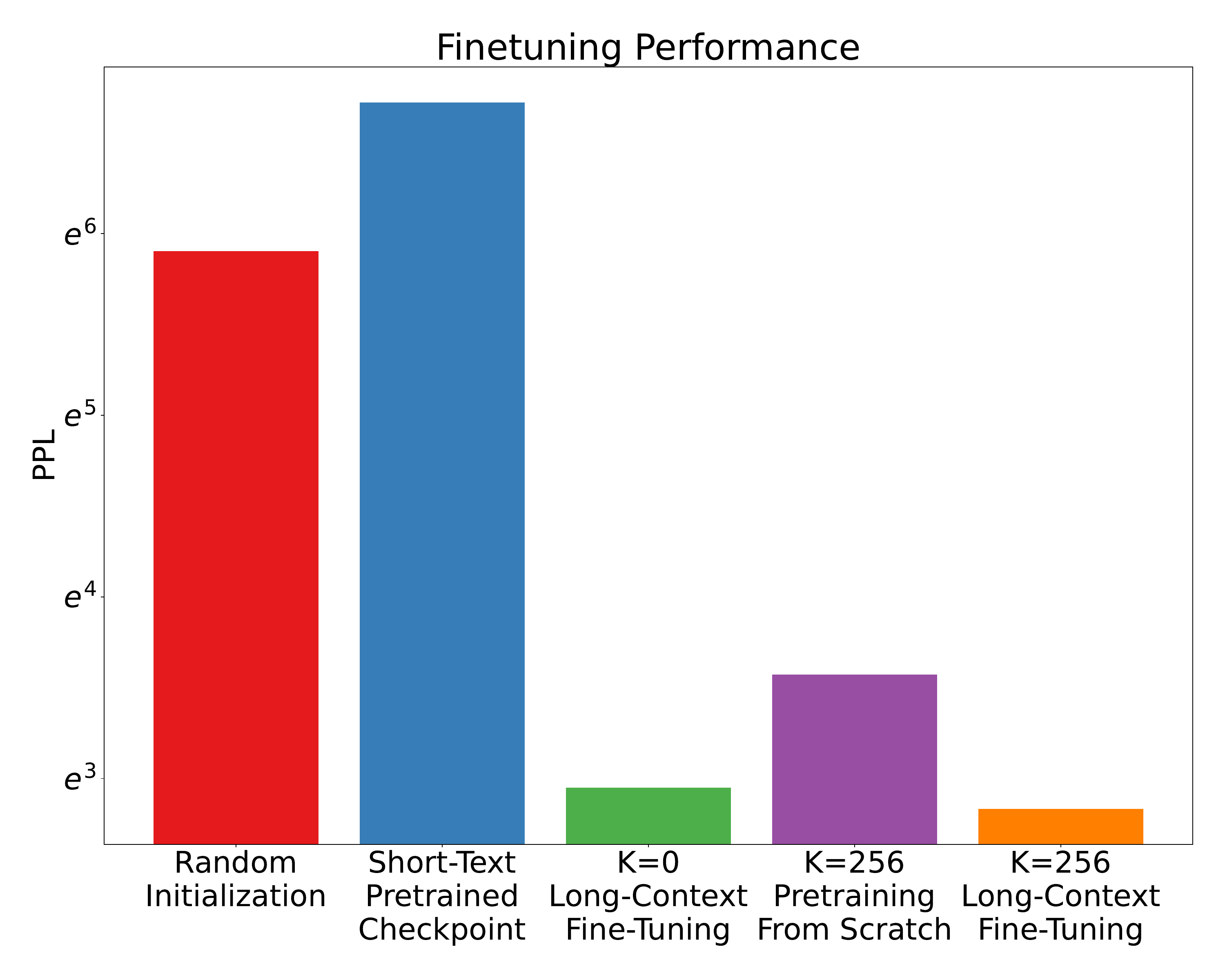}
    \vspace{-5mm}
    \caption{Finetuning Experiments on C4~(4k+) for models pretrained on short-text. We add the best performance achieved in pretraining and a random initialized model performance for reference. We find that short-text pretraining will make model generalize worse even than a randomly initialized model for long-context language modeling, but skipping fine-tuning can help adapt such checkpoints to even perform better than specifically-pretrained long-context checkpoint. }
    \label{fig:finetuning_wo_memory}
    \vspace{-5mm}
    \end{figure}

\paragraph{Using skipping in finetuning can successfully adapt short-text language model a better long-context model.} We present the finetuning results in \cref{fig:finetuning_wo_memory}. Initially, after pretraining on short-text, the model exhibits worse performance (ppl=$831.31$) than a randomly-initialized model (ppl=$366.50$) on long-context language modeling. This indicates that traditional short-text pretraining strategies do not generalize well to long-context scenarios. However, by applying appropriate skipping during finetuning, we can effectively adapt a short-text pretrained model to achieve even better perplexity (ppl=$16.97$, K=$256$) than our best long-context model trained from scratch  (ppl=$35.59$). Moreover, compared to standard finetuning on the long-context corpus (ppl=$19.10$, K=$0$), finetuning with our proposed skipping mechanism  achieves substantial improvements, confirming its effectiveness in enhancing model capabilities in long-context environments.  

Our interpretation is that the ``random chunking and shuffling'' prevalent in traditional short-text pretraining ~\cite{devlin-etal-2019-bert, raffel2020exploring} impairs the model's capability for continuous reading—the input context changes so rapidly that the model lacks the opportunity to capture text coherence effectively. The skipping mechanism can mitigate this by helping to recover long-term dependencies and more effectively generalize the pretrained knowledge in long-context language modeling. 

\begin{figure}[h!]
    \centering
        \centering
        \includegraphics[width=0.93\columnwidth]{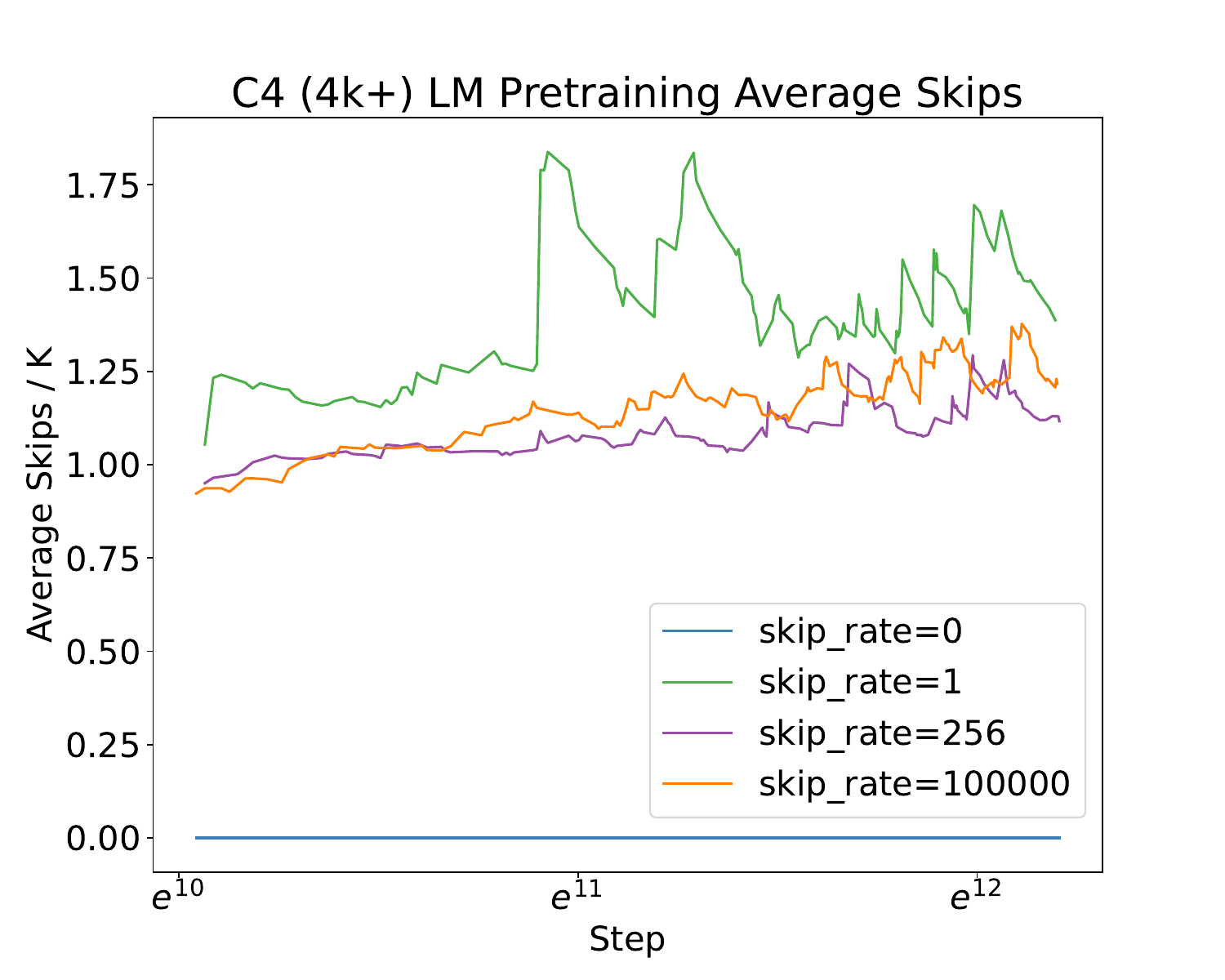}
    \vspace{-5mm}
    \caption{The illustration of the average number of skipped tokens (``Average Skips'') during C4 (4k+) pretraining without using the  memory module. Under different skipping scenarios ($K>0$), the model gradually learns to skip more tokens, indicating it acquires a better understanding of the current long context and becomes more confident in making aggressive skipping decisions.
    }
    \vspace{-5mm}
    \label{fig:avg_moves}
\end{figure}

\paragraph{Model gradually learns to skip more.} In \cref{fig:avg_moves}, we illustrate the average number of skipped tokens during training. For all skipping scenarios ($K > 0$), the average number of skipped tokens increases over time. This indicates the model's improved understanding of the input's long context, leading to more confident and aggressive skipping decisions. By comparing different curves, we observe that with a moderate skipping rate ($K=256$) achieves the smoothest skipping patterns. This may help stabilize training in later phases and achieve better generalization. Although the $K=100,000$ case shows similar patterns, its performance on the pretraining task is worse, indicating that the magnitude of skipping (mainly controlled by skip-rate $K$) is crucial and requires careful tuning.

\paragraph{Choice of pooling does not matter.} We also try several different pooling strategies – ``exponential-decay'', ``only using last token in the current chunk'' and ``average pooling''. It seems that all pooling methods work equally well. 

\begin{figure*}[t!]
\small
    \centering
    \includegraphics[width=0.95\linewidth, trim=0cm 0cm 0cm 0cm, clip]{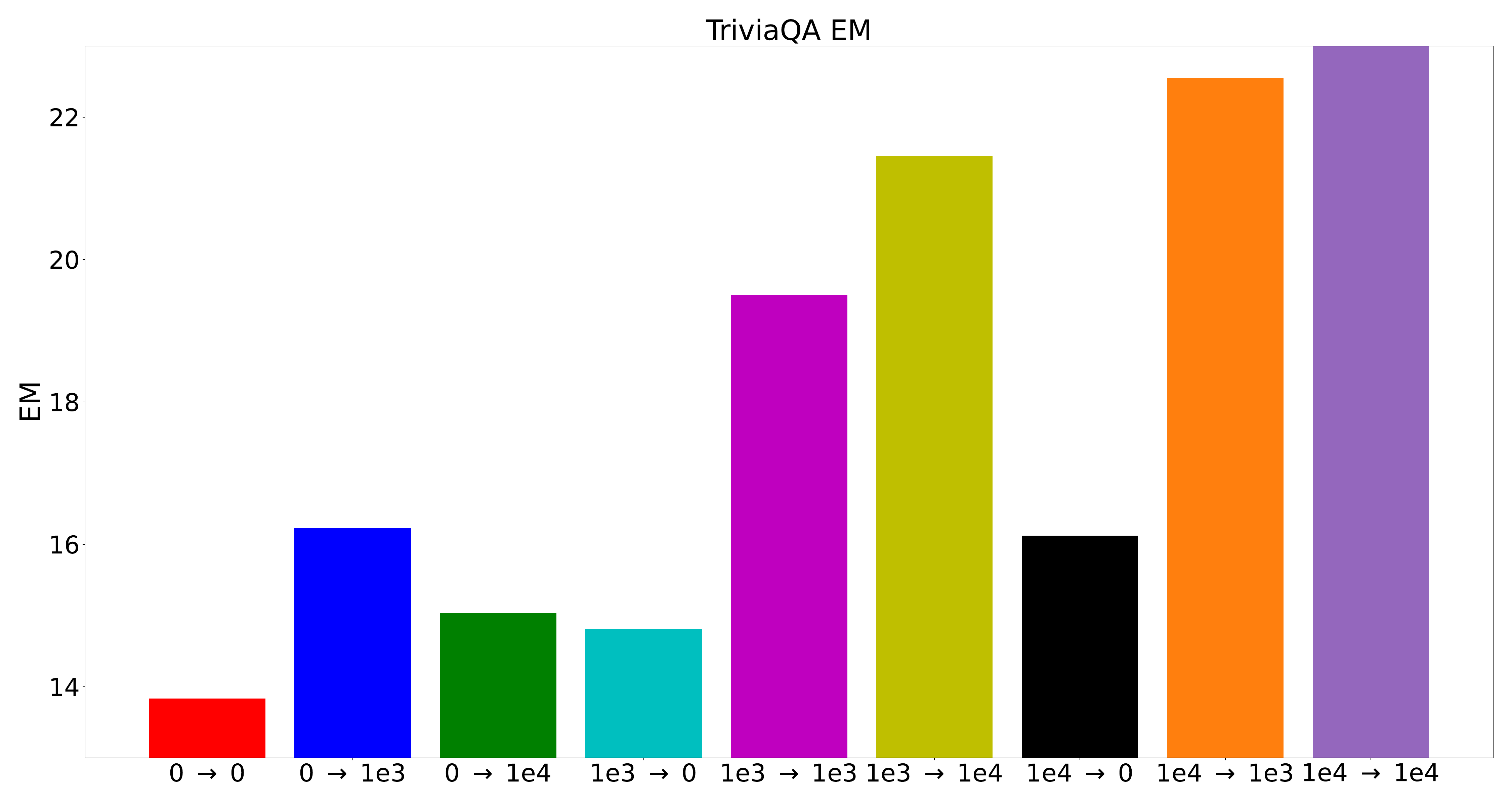}
    \caption{Long-Context Question Answering Experiment results. $x \rightarrow y$ means we train the model with skipping rate $K_{\text{train}}=x$ and evaluated using skipping rate $K_{\text{infer}}=y$. We find that adopting more aggressive skipping strategy helps a lot for improving the model performance. }
    \label{fig:infer_eval_triviaqa}
    \end{figure*}
\begin{table}[h!]
\centering
\resizebox{\columnwidth}{!}{
\begin{tabular}{lcc}
\toprule
Model & PPL & Training Steps \\
\midrule
Transformer w/ Memory~\cite{wu2021memorizing} & 14.97 & 500k \\
Skipping-Pretrained w/o Memory (ours) & 35.59 & 200k \\
Skipping-Pretrained w/ Memory (ours) & \textbf{14.15} & \textbf{130k} \\
\bottomrule
\end{tabular}
}
\caption{Experiments on incorporating memory transformer in pretraining. We adopt the memorizing transformer architectured as implemented in \citet{wu2021memorizing} with 512 context length, 1536 memory size and not using XL-cache. }
\label{tab:w_memory_model_performance}
\end{table}
\paragraph{Combining memory mechanism with skipping brings further improvement.} We find that using skipping in pre-training can benefit transformer pretraining with memory mechanism even if the skipping is not allowed in test time. We adopt one specific memorizing transformer architecture in \cite{wu2021memorizing} (memory-size=1536, context-len=512, no-XL-cache) and perform the same skipping pretraining on C4~(4k+) as above. For fair comparison, we again disable the usage of skipping in test time. The experiment results are shown in \cref{tab:w_memory_model_performance}. We find that our skipping-pretrained memorizing transformer can achieve better performance (ppl=$14.15$) even within first $130$k optimization steps compared with pretrained memorizing transformer reported in \cite{wu2021memorizing} (ppl=$14.97$, $500$k steps), confirming our pretraining methods brings significant performance boost ($-0.82$ ppl) and better learning efficiency (approximately $3.84$ times more efficient).  Compared to our skipping-pretrained w/o memory model (ppl=$35.59$), equipping with the memory mechanism achieves $60\%$ performance gain, demonstrating the tremendous benefit of including memory mechanism in skipping pretraining. This is expected as memory can help track what have read so far explicitly and reduce the memory overload from parameter space in training time. 

\section{Long-Context Question Answering}
\label{sec: long-context qa}
Besides pre-training, we also want to evaluate our skipping-reading models over downstream long-context tasks to verify whether skipping-reading can help. We evaluate our models on TriviaQA dataset~\cite{joshi2017triviaqa} (``RC'' split). We concatenate all retrieved evidences in the input to transform this task to be a relatively challenging long-context question answering task. The average number of tokens for the resulted dataset is $23,706.21$. As inputs now are becoming relatively long, we experiment with more aggressive skipping strategy by setting K$=0, 1000, 10000$ in both training and testing time, leading to $3 \times 3 = 9$ combinations. We random sample $200$ questions for the test set. We use the same transformer architecture as in pretraining experiments. 
For this experiment, we pretrain the model with the memory mechanism introduced above on C4 corpus. 

The experiment results are shown in \cref{fig:infer_eval_triviaqa}. We observe that the best-performing models use a skipping rate of K=$10000$ during both training and testing. Additionally, when the training skipping rate is fixed, using a larger or equal skipping rate during testing consistently yields better results, indicating that the skipping rate can be effectively extrapolated in inference time for more efficient reading. Furthermore, with a fixed inference rate, employing more aggressive training skipping rates consistently leads to improvement. This contrasts with findings in \cref{sec: long-context lm}, where a moderate skipping rate proved more effective. For pretraining, which focuses on next token prediction, the task remains inherently local, and the structure of a long context offers limited benefits. However, in long-context QA, the goal shifts towards enabling the model to handle extremely long documents containing multiple answer-bearing sections and distractors. Skip-reading minimizes redundant engagement with similar relevant passages, thus obviating the need for users to manually filter context. This approach, combined with dynamic loss, achieves effects akin to on-the-fly filtering.

\section{Conclusion}
In this paper, we introduce a random-access reading strategy designed for efficient long-context understanding. The approach primarily utilizes aggregated token-wise loss within each model reading window, enabling a data server to determine the number of tokens to be skipped and execute the skipping process. To enhance decision-making for skipping and provide the model with a richer context, we incorporate an additional memory module. In experiments, we apply our method on Transformer model and evaluate our trained random-access model via  long-context pretraining, fine-tuning, and question answering. Experiment results confirm the significant improvements in performance and efficiency of our proposed simple skipping mechanism over traditional sequential access. Furthermore, these tests demonstrate rapid adaptation of a short-text model to long-context applications, and the mutual benefits of incorporating a memory module and enabling random-access reading. Future work could benefit from exploring a more elaborate skipping mechanism and examining the interplay between the skipping mechanism and the memory module.

\bibliography{example_paper}

\begin{thebibliography}{46}
\providecommand{\natexlab}[1]{#1}
\providecommand{\url}[1]{\texttt{#1}}
\expandafter\ifx\csname urlstyle\endcsname\relax
  \providecommand{\doi}[1]{doi: #1}\else
  \providecommand{\doi}{doi: \begingroup \urlstyle{rm}\Url}\fi

\bibitem[Beltagy et~al.(2020)Beltagy, Peters, and Cohan]{Beltagy2020Longformer}
Beltagy, I., Peters, M.~E., and Cohan, A.
\newblock Longformer: {The} long-document transformer.
\newblock \emph{arXiv:2004.05150}, 2020.

\bibitem[Chen et~al.(2023)Chen, Pasunuru, Weston, and
  Celikyilmaz]{chen2023walking}
Chen, H., Pasunuru, R., Weston, J., and Celikyilmaz, A.
\newblock Walking down the memory maze: {Beyond} context limit through
  interactive reading.
\newblock \emph{arXiv preprint arXiv:2310.05029}, 2023.

\bibitem[Choromanski et~al.(2020)Choromanski, Likhosherstov, Dohan, Song, Gane,
  Sarlos, Hawkins, Davis, Mohiuddin, Kaiser, et~al.]{choromanski2020rethinking}
Choromanski, K.~M., Likhosherstov, V., Dohan, D., Song, X., Gane, A., Sarlos,
  T., Hawkins, P., Davis, J.~Q., Mohiuddin, A., Kaiser, L., et~al.
\newblock Rethinking attention with performers.
\newblock In \emph{International Conference on Learning Representations}, 2020.

\bibitem[Chowdhury \& Caragea(2023)Chowdhury and
  Caragea]{chowdhury2023monotonic}
Chowdhury, J.~R. and Caragea, C.
\newblock Monotonic location attention for length generalization.
\newblock In \emph{International Conference on Machine Learning}. PMLR, 2023.

\bibitem[Deng et~al.(2024)Deng, Gu, Zheng, Chen, Stevens, Wang, Sun, and
  Su]{deng2024mind2web}
Deng, X., Gu, Y., Zheng, B., Chen, S., Stevens, S., Wang, B., Sun, H., and Su,
  Y.
\newblock Mind2web: {Towards} a generalist agent for the web.
\newblock \emph{Advances in Neural Information Processing Systems}, 36, 2024.

\bibitem[Devlin et~al.(2019)Devlin, Chang, Lee, and
  Toutanova]{devlin-etal-2019-bert}
Devlin, J., Chang, M.-W., Lee, K., and Toutanova, K.
\newblock {BERT}: {Pre-training} of deep bidirectional transformers for
  language understanding.
\newblock In Burstein, J., Doran, C., and Solorio, T. (eds.), \emph{Proceedings
  of the 2019 Conference of the North {A}merican Chapter of the Association for
  Computational Linguistics: Human Language Technologies, Volume 1 (Long and
  Short Papers)}, pp.\  4171--4186, Minneapolis, Minnesota, jun 2019.
  Association for Computational Linguistics.
\newblock \doi{10.18653/v1/N19-1423}.
\newblock URL \url{https://aclanthology.org/N19-1423}.

\bibitem[Ding et~al.(2020)Ding, Zhou, Yang, and Tang]{ding2020cogltx}
Ding, M., Zhou, C., Yang, H., and Tang, J.
\newblock Cogltx: {Applying} bert to long texts.
\newblock \emph{Advances in Neural Information Processing Systems},
  33:\penalty0 12792--12804, 2020.

\bibitem[Dong et~al.(2018)Dong, Quirk, and Lapata]{dong2018confidence}
Dong, L., Quirk, C., and Lapata, M.
\newblock Confidence modeling for neural semantic parsing.
\newblock In \emph{Proceedings of the 56th Annual Meeting of the Association
  for Computational Linguistics (Volume 1: Long Papers)}, pp.\  743--753, 2018.

\bibitem[Dong et~al.(2023)Dong, Tang, Li, and Zhao]{dong2023survey}
Dong, Z., Tang, T., Li, L., and Zhao, W.~X.
\newblock A survey on long text modeling with transformers.
\newblock \emph{arXiv preprint arXiv:2302.14502}, 2023.

\bibitem[Fu et~al.(2023)Fu, Epstein, Nguyen, Thomas, Zhang, Dao, Rudra, and
  Re]{fu2023simple}
Fu, D.~Y., Epstein, E.~L., Nguyen, E., Thomas, A.~W., Zhang, M., Dao, T.,
  Rudra, A., and Re, C.
\newblock Simple hardware-efficient long convolutions for sequence modeling.
\newblock In \emph{International Conference on Machine Learning}. PMLR, 2023.

\bibitem[Fu et~al.(2024)Fu, Panda, Niu, Yue, Hajishirzi, Kim, and
  Peng]{fu2024data}
Fu, Y., Panda, R., Niu, X., Yue, X., Hajishirzi, H., Kim, Y., and Peng, H.
\newblock Data engineering for scaling language models to 128k context.
\newblock In \emph{International Conference on Machine Learning}, 2024.

\bibitem[Guo et~al.(2022)Guo, Ainslie, Uthus, Ontanon, Ni, Sung, and
  Yang]{guo2022longt5}
Guo, M., Ainslie, J., Uthus, D.~C., Ontanon, S., Ni, J., Sung, Y.-H., and Yang,
  Y.
\newblock Longt5: {Efficient} text-to-text transformer for long sequences.
\newblock In \emph{Findings of the Association for Computational Linguistics:
  NAACL 2022}, pp.\  724--736, 2022.

\bibitem[Han et~al.(2023)Han, Wang, Xiong, Chen, Ji, and Wang]{han2023lm}
Han, C., Wang, Q., Xiong, W., Chen, Y., Ji, H., and Wang, S.
\newblock Lm-infinite: {Simple} on-the-fly length generalization for large
  language models.
\newblock \emph{arXiv preprint arXiv:2308.16137}, 2023.

\bibitem[Huang et~al.(2023)Huang, Xu, Jiang, Lai, Li, Yao, Chen, Yang, Xin, and
  Ma]{huang2023advancing}
Huang, Y., Xu, J., Jiang, Z., Lai, J., Li, Z., Yao, Y., Chen, T., Yang, L.,
  Xin, Z., and Ma, X.
\newblock Advancing transformer architecture in long-context large language
  models: {A} comprehensive survey.
\newblock \emph{arXiv preprint arXiv:2311.12351}, 2023.

\bibitem[Ivgi et~al.(2023)Ivgi, Shaham, and Berant]{ivgi2023efficient}
Ivgi, M., Shaham, U., and Berant, J.
\newblock Efficient long-text understanding with short-text models.
\newblock \emph{Transactions of the Association for Computational Linguistics},
  11:\penalty0 284--299, 2023.

\bibitem[Jiang et~al.(2021)Jiang, Araki, Ding, and Neubig]{jiang2021can}
Jiang, Z., Araki, J., Ding, H., and Neubig, G.
\newblock How can we know when language models know? on the calibration of
  language models for question answering.
\newblock \emph{Transactions of the Association for Computational Linguistics},
  9:\penalty0 962--977, 2021.

\bibitem[Jimenez et~al.(2023)Jimenez, Yang, Wettig, Yao, Pei, Press, and
  Narasimhan]{jimenez2023swe}
Jimenez, C.~E., Yang, J., Wettig, A., Yao, S., Pei, K., Press, O., and
  Narasimhan, K.~R.
\newblock {SWE-bench}: {Can} language models resolve real-world github issues?
\newblock In \emph{International Conference on Learning Representations}, 2023.

\bibitem[Joshi et~al.(2017)Joshi, Choi, Weld, and
  Zettlemoyer]{joshi2017triviaqa}
Joshi, M., Choi, E., Weld, D.~S., and Zettlemoyer, L.
\newblock {TriviaQA}: {A} large scale distantly supervised challenge dataset
  for reading comprehension.
\newblock In \emph{Proceedings of the 55th Annual Meeting of the Association
  for Computational Linguistics (Volume 1: Long Papers)}, pp.\  1601--1611,
  2017.

\bibitem[Kamath et~al.(2020)Kamath, Jia, and Liang]{kamath2020selective}
Kamath, A., Jia, R., and Liang, P.
\newblock Selective question answering under domain shift.
\newblock In \emph{Proceedings of the 58th Annual Meeting of the Association
  for Computational Linguistics}, pp.\  5684--5696, 2020.

\bibitem[Knuth(1997)]{knuth1997art}
Knuth, D.~E.
\newblock \emph{The art of computer programming}, volume~3.
\newblock Pearson Education, 1997.

\bibitem[Lewis et~al.(2020)Lewis, Perez, Piktus, Petroni, Karpukhin, Goyal,
  K{\"u}ttler, Lewis, Yih, Rockt{\"a}schel, et~al.]{lewis2020retrieval}
Lewis, P., Perez, E., Piktus, A., Petroni, F., Karpukhin, V., Goyal, N.,
  K{\"u}ttler, H., Lewis, M., Yih, W.-t., Rockt{\"a}schel, T., et~al.
\newblock Retrieval-augmented generation for knowledge-intensive nlp tasks.
\newblock \emph{Advances in Neural Information Processing Systems},
  33:\penalty0 9459--9474, 2020.

\bibitem[Liu et~al.(2022)Liu, Ni, Nan, Deb, Zhu, Hassan, and
  Radev]{liu2022leveraging}
Liu, Y., Ni, A., Nan, L., Deb, B., Zhu, C., Hassan, A., and Radev, D.
\newblock Leveraging locality in abstractive text summarization.
\newblock In \emph{Proceedings of the 2022 Conference on Empirical Methods in
  Natural Language Processing}, pp.\  6081--6093, 2022.

\bibitem[Liu et~al.(2023)Liu, Wang, Dao, Zhou, Yuan, Song, Shrivastava, Zhang,
  Tian, Re, et~al.]{liu2023deja}
Liu, Z., Wang, J., Dao, T., Zhou, T., Yuan, B., Song, Z., Shrivastava, A.,
  Zhang, C., Tian, Y., Re, C., et~al.
\newblock Deja vu: {Contextual} sparsity for efficient {LLMs} at inference
  time.
\newblock In \emph{International Conference on Machine Learning}, pp.\
  22137--22176. PMLR, 2023.
\newblock URL \url{https://proceedings.mlr.press/v202/liu23am.html}.

\bibitem[Ma et~al.(2021)Ma, Kong, Wang, Zhou, May, Ma, and
  Zettlemoyer]{ma2021luna}
Ma, X., Kong, X., Wang, S., Zhou, C., May, J., Ma, H., and Zettlemoyer, L.
\newblock Luna: {Linear} unified nested attention.
\newblock \emph{Advances in Neural Information Processing Systems},
  34:\penalty0 2441--2453, 2021.

\bibitem[Mohtashami \& Jaggi(2023)Mohtashami and Jaggi]{mohtashami2023landmark}
Mohtashami, A. and Jaggi, M.
\newblock Landmark attention: {Random-access} infinite context length for
  transformers.
\newblock \emph{Advances in Neural Information Processing Systems}, 2023.

\bibitem[Mohtashami \& Jaggi(2024)Mohtashami and Jaggi]{mohtashami2024random}
Mohtashami, A. and Jaggi, M.
\newblock Random-access infinite context length for transformers.
\newblock \emph{Advances in Neural Information Processing Systems}, 36, 2024.

\bibitem[Mou et~al.(2021)Mou, Yang, Yu, Yao, Guo, Potdar, and
  Su]{mou2021narrative}
Mou, X., Yang, C., Yu, M., Yao, B., Guo, X., Potdar, S., and Su, H.
\newblock Narrative question answering with cutting-edge open-domain qa
  techniques: {A} comprehensive study.
\newblock \emph{Transactions of the Association for Computational Linguistics},
  9:\penalty0 1032--1046, 2021.

\bibitem[Nguyen et~al.(2022)Nguyen, Pham, Nguyen, Nguyen, Osher, and
  Ho]{nguyen2022fourierformer}
Nguyen, T., Pham, M., Nguyen, T., Nguyen, K., Osher, S., and Ho, N.
\newblock Fourierformer: {Transformer} meets generalized fourier integral
  theorem.
\newblock \emph{Advances in Neural Information Processing Systems},
  35:\penalty0 29319--29335, 2022.

\bibitem[Paris et~al.(1991)Paris, Wasik, and Turner]{paris1991development}
Paris, S.~G., Wasik, B., and Turner, J.~C.
\newblock The development of strategic readers.
\newblock 1991.

\bibitem[Peng et~al.(2023)Peng, Quesnelle, Fan, and Shippole]{peng2023yarn}
Peng, B., Quesnelle, J., Fan, H., and Shippole, E.
\newblock Yarn: {Efficient} context window extension of large language models.
\newblock \emph{arXiv preprint arXiv:2309.00071}, 2023.

\bibitem[Peng et~al.(2020)Peng, Pappas, Yogatama, Schwartz, Smith, and
  Kong]{peng2020random}
Peng, H., Pappas, N., Yogatama, D., Schwartz, R., Smith, N., and Kong, L.
\newblock Random feature attention.
\newblock In \emph{International Conference on Learning Representations}, 2020.

\bibitem[Pressley \& Afflerbach(2012)Pressley and
  Afflerbach]{pressley2012verbal}
Pressley, M. and Afflerbach, P.
\newblock \emph{Verbal protocols of reading: {The} nature of constructively
  responsive reading}.
\newblock Routledge, 2012.

\bibitem[Qiu et~al.(2020)Qiu, Ma, Levy, Yih, Wang, and Tang]{qiu2020blockwise}
Qiu, J., Ma, H., Levy, O., Yih, W.-t., Wang, S., and Tang, J.
\newblock Blockwise self-attention for long document understanding.
\newblock In \emph{Findings of the Association for Computational Linguistics:
  EMNLP 2020}, pp.\  2555--2565, 2020.

\bibitem[Raffel et~al.(2020)Raffel, Shazeer, Roberts, Lee, Narang, Matena,
  Zhou, Li, and Liu]{raffel2020exploring}
Raffel, C., Shazeer, N., Roberts, A., Lee, K., Narang, S., Matena, M., Zhou,
  Y., Li, W., and Liu, P.~J.
\newblock Exploring the limits of transfer learning with a unified text-to-text
  transformer.
\newblock \emph{Journal of machine learning research}, 21\penalty0
  (140):\penalty0 1--67, 2020.

\bibitem[Reid et~al.(2024)Reid, Savinov, Teplyashin, Lepikhin, Lillicrap,
  Alayrac, Soricut, Lazaridou, Firat, Schrittwieser, et~al.]{reid2024gemini}
Reid, M., Savinov, N., Teplyashin, D., Lepikhin, D., Lillicrap, T., Alayrac,
  J.-b., Soricut, R., Lazaridou, A., Firat, O., Schrittwieser, J., et~al.
\newblock Gemini 1.5: {Unlocking} multimodal understanding across millions of
  tokens of context.
\newblock \emph{arXiv preprint arXiv:2403.05530}, 2024.

\bibitem[Shi et~al.(2023)Shi, Min, Yasunaga, Seo, James, Lewis, Zettlemoyer,
  and Yih]{shi2023replug}
Shi, W., Min, S., Yasunaga, M., Seo, M., James, R., Lewis, M., Zettlemoyer, L.,
  and Yih, W.-t.
\newblock Replug: {Retrieval-augmented} black-box language models.
\newblock \emph{arXiv preprint arXiv:2301.12652}, 2023.

\bibitem[Su et~al.(2024)Su, Ahmed, Lu, Pan, Bo, and Liu]{su2024roformer}
Su, J., Ahmed, M., Lu, Y., Pan, S., Bo, W., and Liu, Y.
\newblock Roformer: {Enhanced} transformer with rotary position embedding.
\newblock \emph{Neurocomputing}, 568:\penalty0 127063, 2024.

\bibitem[Tay et~al.(2020{\natexlab{a}})Tay, Bahri, Yang, Metzler, and
  Juan]{tay2020sparse}
Tay, Y., Bahri, D., Yang, L., Metzler, D., and Juan, D.-C.
\newblock Sparse sinkhorn attention.
\newblock In \emph{International Conference on Machine Learning}, pp.\
  9438--9447. PMLR, 2020{\natexlab{a}}.

\bibitem[Tay et~al.(2020{\natexlab{b}})Tay, Dehghani, Abnar, Shen, Bahri, Pham,
  Rao, Yang, Ruder, and Metzler]{tay2020long}
Tay, Y., Dehghani, M., Abnar, S., Shen, Y., Bahri, D., Pham, P., Rao, J., Yang,
  L., Ruder, S., and Metzler, D.
\newblock Long range arena: {A} benchmark for efficient transformers.
\newblock In \emph{International Conference on Learning Representations},
  2020{\natexlab{b}}.

\bibitem[Wu et~al.(2021)Wu, Rabe, Hutchins, and Szegedy]{wu2021memorizing}
Wu, Y., Rabe, M.~N., Hutchins, D., and Szegedy, C.
\newblock Memorizing transformers.
\newblock In \emph{International Conference on Learning Representations}, 2021.

\bibitem[Xiao et~al.(2024)Xiao, Tian, Chen, Han, and Lewis]{xiao2023efficient}
Xiao, G., Tian, Y., Chen, B., Han, S., and Lewis, M.
\newblock Efficient streaming language models with attention sinks.
\newblock In \emph{International Conference on Learning Representations}, 2024.

\bibitem[Xiong et~al.(2023)Xiong, Liu, Molybog, Zhang, Bhargava, Hou, Martin,
  Rungta, Sankararaman, Oguz, et~al.]{xiong2023effective}
Xiong, W., Liu, J., Molybog, I., Zhang, H., Bhargava, P., Hou, R., Martin, L.,
  Rungta, R., Sankararaman, K.~A., Oguz, B., et~al.
\newblock Effective long-context scaling of foundation models.
\newblock \emph{arXiv preprint arXiv:2309.16039}, 2023.

\bibitem[Yang \& Ettinger(2023)Yang and Ettinger]{yang2023can}
Yang, C. and Ettinger, A.
\newblock Can you follow me? testing situational understanding for {ChatGPT}.
\newblock In \emph{Proceedings of the 2023 Conference on Empirical Methods in
  Natural Language Processing}, pp.\  6385--6398, 2023.

\bibitem[Yang \& Hua(2024)Yang and Hua]{yang2024attendre}
Yang, Z. and Hua, N.
\newblock Attendre: {Wait} to attend by retrieval with evicted queries in
  memory-based transformers for long context processing.
\newblock \emph{arXiv preprint arXiv:2401.04881}, 2024.

\bibitem[Zhang et~al.(2022)Zhang, Ni, Mao, Wu, Zhu, Deb, Awadallah, Radev, and
  Zhang]{zhang2022summn}
Zhang, Y., Ni, A., Mao, Z., Wu, C.~H., Zhu, C., Deb, B., Awadallah, A., Radev,
  D., and Zhang, R.
\newblock Summn: {A} multi-stage summarization framework for long input
  dialogues and documents.
\newblock In \emph{Proceedings of the 60th Annual Meeting of the Association
  for Computational Linguistics (Volume 1: Long Papers)}, pp.\  1592--1604,
  2022.

\bibitem[Zhou et~al.(2023)Zhou, Xu, Zhu, Zhou, Lo, Sridhar, Cheng, Ou, Bisk,
  Fried, et~al.]{zhou2023webarena}
Zhou, S., Xu, F.~F., Zhu, H., Zhou, X., Lo, R., Sridhar, A., Cheng, X., Ou, T.,
  Bisk, Y., Fried, D., et~al.
\newblock Webarena: {A} realistic web environment for building autonomous
  agents.
\newblock In \emph{International Conference on Learning Representations}, 2023.

\end{thebibliography}
\bibliographystyle{icml2023}

\newpage
\appendix
\onecolumn

\end{document}